\documentclass[journal]{IEEEtran}
\usepackage{pdfpages}
\usepackage{float}

\usepackage{cite}

\ifCLASSINFOpdf
\else
\fi
\usepackage{amsmath}
\usepackage{algorithmic}
\usepackage{amsmath,amssymb,amsfonts}
\usepackage[british]{babel}

\usepackage{booktabs}
\usepackage{multirow}
\usepackage{array}

\begin{document}
%
\title{PolSAR Image Classification Based on Robust Low-Rank Feature Extraction and Markov Random Field}
%
%
%

\author{
Haixia~Bi, Jing~Yao, Zhiqiang Wei, Danfeng~Hong,~\IEEEmembership{Member,~IEEE,} and Jocelyn Chanussot,~\IEEEmembership{Fellow,~IEEE}
\thanks{
H.~Bi is with Faculty of Engineering, University of Bristol, Bristol
BS8 1UB, United Kingdom. (e-mail: haixia.bi@bristol.ac.uk)

J.~Yao is with the School of Mathematics and Statistics, Xi'an Jiaotong University, 710049 Xi'an, China. (e-mail: jasonyao@stu.xjtu.edu.cn)

Z.~Wei is with the Xi'an Electronics and Engineering Institute,
710100 Xi'an, China. (e-mail: zqwei@fudan.edu.cn)

D. Hong is with the Remote Sensing Technology Institute (IMF), German Aerospace Center (DLR), 82234 Wessling, Germany. (e-mail: danfeng.hong@dlr.de)

J. Chanussot is with the Aerospace Information Research Institute, Chinese Academy of Sciences, 100094 Beijing, China. (e-mail: jocelyn@hi.is)
}}

%
%

\markboth{SUBMISSION TO IEEE GEOSCIENCE AND REMOTE SENSING LETTERS}%
{Shell \MakeLowercase{\textit{et al.}}: Bare Demo of IEEEtran.cls for IEEE Journals}
%



\maketitle
\begin{abstract}
Polarimetric synthetic aperture radar (PolSAR) image classification has been investigated vigorously in various remote sensing applications. However, it is still a challenging task nowadays. One significant barrier lies in
the speckle effect embedded in the PolSAR imaging process,
which greatly degrades the quality of the images
and further complicates the classification. To this end, we present a novel PolSAR image classification method, which removes speckle noise via low-rank (LR) feature extraction and enforces smoothness priors via Markov random field (MRF). Specifically, we employ the mixture of Gaussian-based robust LR matrix factorization to simultaneously extract discriminative features and remove complex noises. Then, a classification map is obtained by applying convolutional neural network with data augmentation on the extracted features, where local consistency is implicitly involved, and the insufficient label issue is alleviated. Finally, we refine the classification map by MRF to enforce contextual smoothness. We conduct experiments on two benchmark PolSAR datasets. Experimental results indicate that the proposed method achieves promising classification performance and preferable spatial consistency.
\end{abstract}

\IEEEpeerreviewmaketitle

\section{Introduction}

\IEEEPARstart{P}{olarimetric} synthetic aperture radar (PolSAR) is
an advantageous earth observation technique nowadays.
It captures abundant and high-resolution earth surface information
by transmitting and receiving radar signals in different polarimetric ways.
Earth terrains can be retrieved
via effective PolSAR image classification techniques,
which exposes huge theoretical and application significance
in civil and military fields.
PolSAR image classification has been delved into extensively
in recent years.

However, despite the rapid development,
there are still some challenges confronting the
PolSAR image classification task,
wherein the presence of speckle noise
is one of the most significant ones.
Researchers have put forward lots of methods
to handle the adverse impact of various complex noises \cite{he2013texture,masjedi2015classification,zhou2016polarimetric,bi2018graph,bi2019active,hong2019augmented,yao2019nonconvex,bi2017polsar,bi2017unsupervised,zhang2017complex,gao2020spectral}. One conventional way is to design handcrafted features
using sliding window techniques.
Wavelet technique \cite{he2013texture} was leveraged
to extract polarization wavelet features
in the space of polarization states.
Literature \cite{masjedi2015classification} proposed to use Gabor filtering and wavelet analysis to obtain contextual features. A more heuristic way is to extract features
based on learning-based methods, e.g., convolutional neural network (CNN) \cite{zhou2016polarimetric,chen2018polsar,hong2020graph}. They have been employed to automatically extract data-driven polarimetric features, and were further combined with semi-supervised learning and active learning  in literature \cite{bi2018graph,bi2019active}.
A complex-valued CNN was applied in PolSAR image classification in \cite{zhang2017complex}.
An alternative strategy is to
encourage local consistency on pixel labels.
Typical methods include the over-segmentation technique \cite{hong2020invariant}
used in preprocessing stage
and Markov random field (MRF)-based optimization employed
as a post-processing process \cite{bi2017polsar,bi2017unsupervised}.

\begin{figure*}[!t]
\vspace{-0.2cm}
\begin{center}
\includegraphics[height=2.4cm,width=15cm]{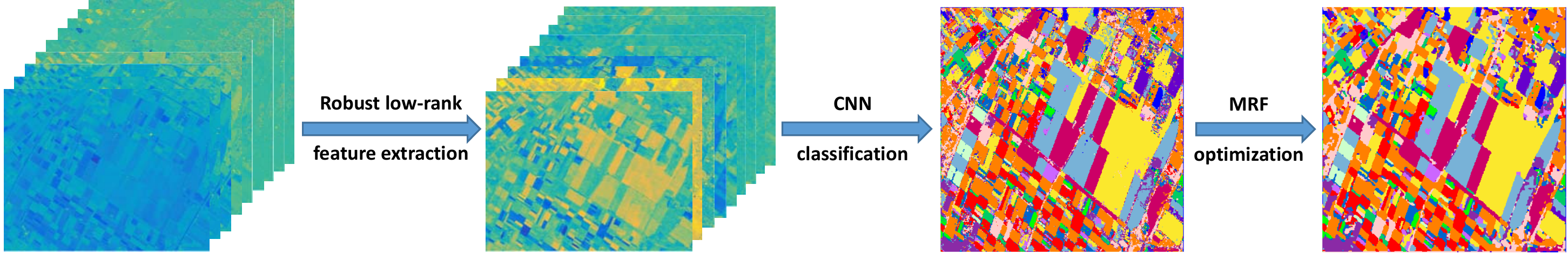}
\vspace{-0.2cm}
\caption{\small Illustrating the PolSAR image classification
method via robust low-rank feature extraction and Markov random field.}
\label{fig.framework}
\end{center}
\vspace{-0.6cm}
\end{figure*}

Although the previous approaches achieved good results on
mitigating the performance degradation caused by speckle noise,
they do not explicitly consider removing
noise while extracting features.
To solve this problem from the source,
we propose to simultaneously perform noise removal
and feature extraction.
Specifically,
we leverage the mixture of Gaussian (MoG)-based
robust low-rank matrix factorization (RLRMF) to eliminate noise disturbances
and draw compact and robust polarimetric features,
where the universal approximation power of
MoG model is exploited
to characterize the PolSAR noise
with complex and heterogenous distributions.
With the extracted RLRMF features,
a classification map is then obtained by applying
data augmentation incorporated CNN classifier on them,
where local consistency is implicitly enforced
by the convolution operations.
Finally, based on the output probabilities of the CNN classifier,
the classification map is refined
by solving an MRF optimization problem
to further enforce spatial smoothness.

This paper presents a CNN-based PolSAR image classification method incorporating RLRMF and MRF to suppress speckle noises, further promoting the classification performance.
The main contributions of this letter are two-fold:
\begin{enumerate}
\item The speckle noise is suppressed
in three different dimensions, i.e., mainly reduced via RLRMF-based noise removal and feature extraction,
suppressed by CNN-based classification, and MRF prior-based post-processing. To our knowledge, this is the first work that considers suppressing complex noises from diverse views and extracting discriminative features for classification.

\item
RLRMF features with data augmentation can provide robust and discriminative information for CNN classifier, which effectively alleviates the insufficient label problem.
Beyond the traditional MRF model that only enforces label smoothness priors, our MRF not only encourages the contextual consistency, but also enforces the alignment of label boundaries with image edges.


\end{enumerate}


\section{Proposed Method}

\subsection{Problem Formulation}
The input polarimetric data for each pixel
is a $3 \times 3$ complex conjugate symmetric coherency matrix $\mathbf{T}$.
For convenience, $\mathbf{T}$
can be vectorized to a $9 \times 1$ real vector as

\vspace{-3mm}
\begin{eqnarray}\label{Fea9D}
\begin{split}
&[\mathbf{T}_{11}, \mathbf{T}_{22}, \mathbf{T}_{33}, \Re({\mathbf{T}}_{12}), \Im({\mathbf{T}}_{12}), \\ &\rm{\quad\;}\Re({\mathbf{T}}_{13}),
\Im({\mathbf{T}}_{13}),\Re({\mathbf{T}}_{23}), \Im({\mathbf{T}}_{23})]^\emph{T},
\end{split}
\end{eqnarray}
where $\Re{(\cdot)}$ and $\Im{(\cdot)}$
extract the real and imaginary components of
the cross-polarization items.
Superscript $^T$ denotes the transpose operation.
Each vector dimension in Eq.~(\ref{Fea9D})
is normalized to [0,1] before feeding into the proposed method.

Then for a given PolSAR image,
the raw PolSAR data set can be represented by
$\mathbf{X}=\{\mathbf{x}_i \in \mathbb{R}^d\}_{i=1}^N$,
$N=H\times W$. Here, $H$ and $W$ are height and width
of the image, $d$ is the dimension of the raw feature,
which takes the value of 9 as defined in Eq.~(1).
We define the labeled training samples as ${(\mathbf{x}_i,\mathbf{y}_i)}_{i=1}^{n}$,
where $n \ll N$, $\mathbf{y}_i \in \{1, \cdots, C \}$,
$C$ is the total number of classes.
Then the proposed method is designed to assign class label $\mathbf{y}_i$
to each pixel $i$.
We further denote $\mathbf{Y}=\{\mathbf{y}_i\}_{i=1}^{N}$ in the following sections.
In addition, we denote $\mathbf{F}_{ij}$ as the pixel vector at
location $(i, j)$, and $\mathbf{S}_{ij}=\cup_{\{(u,v):|u-i| \leq \lfloor \frac{s}{2} \rfloor,
|v-j| \leq \lfloor \frac{s}{2} \rfloor \}}\mathbf{F}_{uv}$ as the neighborhood of $\mathbf{F}_{ij}$
in the spatial dimension by using a sliding window box
with size $s \times s$. Here, $s$ is the window size,
and $\lfloor \frac{s}{2} \rfloor$ is the rounded down integer
of number $\frac{s}{2}$.

Taking the Flevoland area data set as example, Fig.~\ref{fig.framework}
illustrates the pipeline of our proposed method.
The proposed RLRMF feature extraction is first implemented
to remove complex noise while extracting features (Section II-B).
We then train a CNN classifier to obtain
a classification map,
and the classification map is finally refined
by solving an MRF prior-based optimization problem. (Section II-C).

\subsection{Feature Extraction Using RLRMF}
Since neighboring pixels in PolSAR images are prone to
have high affinities, i.e., PolSAR data has low-rank characteristics,
RLRMF is thus an effective tool to
capture the global structures of the data and remove noise \cite{meng2013robust}.
For each spatial neighborhood matrix
$\mathbf{S} \in \mathbb{R}^{d\times s^2}$
(we omit subscript $ij$ for simplicity),
the RLRMF problem can be formulated as,

\vspace{-3mm}
\begin{eqnarray} \label{LRMF}
\min_{\mathbf{U},\mathbf{V}} \left\| \mathbf{S}-\mathbf{U}\mathbf{V}^T \right\|,
\end{eqnarray}
where $\mathbf{U}\in \mathbb{R}^{d \times r}$ and $\mathbf{V}\in \mathbb{R}^{s^2 \times r}$
are matrices with low-rank property $(r < \min(s^2, d))$.
$\left\| \cdot \right\|$ represents the noise measure of PolSAR data.
After obtaining an optimal solution $(\mathbf{U}^*,\mathbf{V}^*)$, $\mathbf{S}$ can be recovered by the product $\mathbf{U}^*\mathbf{V}^{*T}$.

Due to the complexity and heterogeneity of polarimetric noise,
MoG assumption is used to model the unknown noise distributions.
We denote the noise term as $\textbf{E}$ with entries $\epsilon_{ij}$,
each of which is a sample from MoG distribution
$p(\epsilon)\sim \sum_{k=1}^{K}\pi_k \mathcal{N}(\epsilon;0,\sigma_k^2)$,
where $\mathcal{N}(\epsilon;0,\sigma^2)$ indicates the Gaussian distribution
with mean 0 and variance $\sigma^2$.
Here, $\pi_k>0$, and $\sum_{k=1}^{K}\pi_k=1$, where $K$
is the number of mixed components.
Under maximum likelihood estimation framework,
the log-likelihood of noise term $\textbf{E}$
is written as,

\vspace{-3mm}
\begin{eqnarray} \label{LH}
P(\mathbf{E})=\sum_{i=1}^{d}\sum_{j=1}^{s^2}\text{log}\sum_{k=1}^{K}\pi_k
\mathcal{N}(\epsilon_{ij}|0,\sigma_k^2).
\end{eqnarray}

Since each element $\mathbf{s}_{ij}$ in $\mathbf{S}$ satisfies
$\epsilon_{ij}=\mathbf{s}_{ij}-\mathbf{u}_i^T\mathbf{v}_j$,
where $\mathbf{u_i}$ and $\mathbf{v_j}$
are the $i$th and $j$th column vectors of
$\mathbf{U}$ and $\mathbf{V}$ respectively,
let $\Pi=\{\pi_1, \pi_2,\cdots, \pi_K\}$,
$\Sigma=\{\sigma_1, \sigma_2,\cdots, \sigma_K\}$,
and $\Theta=\{\mathbf{U},\mathbf{V},\Pi,\Sigma\}$,
our objective is to maximize
the log-likelihood function as,

\vspace{-2mm}
\begin{equation}\label{eqn.likelihood}
    \begin{aligned}
&\max_{\Theta} \mathcal{L}(\Theta)
&=\sum_{i=1}^{d}\sum_{j=1}^{s^2}\text{log}\sum_{k=1}^{K}\pi_k
\mathcal{N}(\mathbf{s}_{ij}-\mathbf{u}_i^T\mathbf{v}_j|0,\sigma_k^2).
\end{aligned}
\end{equation}

The Expectation-Maximization (EM) algorithm
is employed to solve the model defined in Eq.~\ref{eqn.likelihood}.
That is, the parameters
are estimated by iteratively conducting \emph{E-step}, i.e.,
calculating posterior probabilities of all Gaussian components,
and \emph{M-step}, i.e., re-estimating the parameters
by maximizing the Q-function which is the upper-bound
of the original likelihood.

\textbf{E-Step:}
We assume a latent variable $z_{ijk}$ in the model with
$z_{ijk} \in \{0, 1\}$ and $\sum_{k=1}^K=1$,
indicating the assignment of the noise
$\epsilon_{ij}$ to a $k$th component of the mixture.
Then, the posterior possibility of mixture $k$ for
generating the noise of $\mathbf{s}_{ij}$
can be calculated by,

\vspace{-3mm}
\begin{eqnarray}\label{eqn.estep}
\begin{split}
E(z_{ijk})=\gamma_{ijk}=\frac{\pi_k \mathcal{N}(\mathbf{s}_{ij}|\mathbf{u}_i^T\mathbf{v}_j,\sigma_k^2)}
{\sum_{k=1}^K \pi_k \mathcal{N}(\mathbf{s}_{ij}|\mathbf{u}_i^T\mathbf{v}_j,\sigma_k^2)}.
\end{split}
\end{eqnarray}

\textbf{M-Step:}
M step maximizes the Q-function constructed
from the posterior possibility in E-step,

\vspace{-3mm}
\begin{eqnarray}\label{eqn.Qfunc}
E_{\mathbf{Z}}p(\mathbf{S},\mathbf{Z}|\Theta)=
\sum_{i,j,k}
\gamma_{ijk} (\text{log}\frac{\pi_k}{\sqrt{2\pi}\sigma_k}-
\frac{(\mathbf{s}_{ij}-\mathbf{u}_i^T\mathbf{v}_j)^2}{2\pi\sigma_k^2}).
\end{eqnarray}

Then we solve this optimization problem by alternatively
updating the MoG parameters $\mathbf{U},\mathbf{V}$ and the factorized
matrices $\Pi,\Sigma$ as follows:

(1) Update $\Pi,\Sigma$:
Closed-form solutions for the MoG parameters can be given by,

\vspace{-3mm}
\begin{eqnarray} \label{eqn.MoG}
\sigma_k^2=\frac{1}{\sum_{i,j} \gamma_{ijk}}\sum_{i=1}^d\sum_{j=1}^{s^2}
\gamma_{ijk}(\mathbf{S}_{ij}-\mathbf{u}_i^T\mathbf{v}_j)^2.
\end{eqnarray}

(2) Update $\mathbf{U},\mathbf{V}$:
By reformulating the objective function in Eq.~\ref{eqn.Qfunc} w.r.t $\mathbf{U}$ and $\mathbf{V}$, we have below optimization problem,

\vspace{-3mm}
\begin{eqnarray} \label{eqn.UV}
\min_{\mathbf{U},\mathbf{V}} \left\| \mathbf{W}\odot(\mathbf{S}-\mathbf{U}\mathbf{V}^T) \right\|_F^2,
\end{eqnarray}
where the element $w_{ij}$ of $\mathbf{W}$ is $\sum_{k=1}^K \frac{\gamma_{ijk}}{2\pi\sigma_k}$.
Problem (\ref{eqn.UV}) can be solved by \cite{hong2019learnable}.

With the solution $(\mathbf{U}^*,\mathbf{V}^*)$
obtained using the EM algorithm, the spatial
neighborhood $\mathbf{S}$ can be recovered by
$\hat{\mathbf{S}}=\mathbf{U}^*\mathbf{V}^{*T}$ .
Then, the extracted RLRMF feature $\hat{\mathbf{F}}$
is used as input in the consequent classification process
instead of the original feature ${\mathbf{F}}$.

\subsection{CNN-MRF Classification}

Our proposed CNN-MRF PolSAR classification method is composed of two steps: CNN classification and MRF prior-based optimization.

\subsubsection{CNN Classification}
CNN is leveraged to predict class label
based on the RLRMF features.
We adopt a CNN architecture \cite{rasti2020feature} which consists of two groups of convolutional layer and pooling layer,
followed by two fully connected layers and an output softmax layer.
Batch normalization is performed after the two convolutional layers
and the first fully connected layer.
We exploit data augmentation technique
following ~\cite{bi2019active} to enlarge the training set,
alleviating the reliance of CNN on a large amount of labels.
Let $\Phi$ denote the combination of the filters, weight matrices
and biases of the CNN,
the loss function of the CNN is defined as
\begin{eqnarray} \label{Optimfun1}
\begin{split}
E_d=-\sum_{i=1}^{n}\sum_{j=1}^C 1\{\mathbf{y}_i=j\}
\log P(\mathbf{y}_i=j|{F_\Phi},\hat{\mathbf{f}}_i),    \\
\end{split}
\end{eqnarray}
where $P(\mathbf{y}_i=j|{F_{\Phi}},\hat{\mathbf{f}}_i)$ denotes the probability
of $\hat{\mathbf{f}}_i$ having label $j$.
We first train a CNN by minimizing the loss function in Eq.~(\ref{Optimfun1}).
Then the learned classifier $F_{\Phi}$ is applied on the whole data set
to obtain class probabilities for the MRF optimization.

\subsubsection{MRF Prior-based Optimization}

This step is used to predict the final labels $\mathbf{Y}$
under the constraint of MRF priors where
the smoothness of estimated label map
and the alignment of label boundaries with image edges are enforced.
This task can be expressed as the following optimization problem,

\vspace{-0.3mm}
\begin{eqnarray} \label{eqn.smooth}
\mathbf{Y}=\mathop{\arg\min}\limits_{\mathbf{y} \in \{1,...,C \}}(-\sum_{i=1}^{N}\log{P(\mathbf{y}_i})+
  \alpha_s\sum_{i=1}^N\sum_{j\in \mathcal{N}(i)}\mathcal{S}_{ij}).
\end{eqnarray}
The first part of the optimization objective
is the CNN output probability,
and the second part is the label smoothness term
with $\mathcal{S}_{ij}=|\mathbf{y}_i-\mathbf{y}_j|\exp(-\frac{\|\mathbf{z}_{i}-\mathbf{z}_{j}\|_2^2}{2\sigma})$, $\alpha_s$ as the label smoothness factor,
and $\mathcal{N}(i)$ as the neighboring pixel set of pixel $i$.
$\mathbf{z}_{i}$ should take features
significantly changing values across the image edges.
Due to the proper delineation of homogeneous areas and edges, Pauli matrix components are taken as $\mathbf{z}_{i}$,  where $\delta$ indicates the mean squared distance between features of two adjacent pixels. Solving Eq.~\ref{eqn.smooth} is a combinatorial optimization problem. Belief propagation \cite{tappen2003comparison} optimization algorithm is employed here due to its quick convergence.

\section{Experiments}
In this section, we validate our proposed method
on two real PolSAR data sets.
We first introduce the experimental data sets and parameter settings
in Section III-A.
Next, the experimental results are reported in Section III-B, where comparison is performed between the proposed CNN-RLRMF-MRF method (denoted as Ours) and one traditional method SVM~\cite{hong2019cospace},
and three state-of-the-art competitors, including CNN~\cite{zhou2016polarimetric}, CNN-MRF~\cite{cao2019spectral} and complex-valued CNN (CV-CNN)~\cite{zhang2017complex}.
We conduct ablation study in Section III-C, to demonstrate the validity of the key components of the proposed method.

\begin{table}[!t]
    \begin{center}
    \caption{Used labels on two data sets of different methods}
    \vspace{-0.1cm}
        \begin{tabular} {c|c|c}
            \hline
            \hline
            {Method} & Flevoland & Oberpfaffenhofen \\
            \hline
            {SVM \cite{hong2019cospace}} & {13535 (10\%)} & {137429 (10\%)} \\
            {CNN \cite{zhou2016polarimetric}} & {13535 (10\%)} & {137429 (10\%)} \\
            {CNN-MRF \cite{cao2019spectral}} & {13535 (10\%)} & {137429 (10\%)} \\
            {CV-CNN \cite{zhang2017complex}} & {11900 (8.8\%)} & {13743 (1\%)} \\
            {Ours} & {2707 (2\%)} & ~~{6872 (0.5\%)} \\
            \hline
            \hline
        \end{tabular}
    \vspace{-0.1cm}
    \label{tbl.usedlabels}
    \end{center}
\end{table}

\begin{table*}[!t]
	\caption{CAs (\%) and OAs (\%) of Flevoland area data set with different methods}
    \vspace{-0.1cm}
	\begin{center}
		\begin{tabular}{c|cccccccccccccc|c}
			\hline	\hline
    		Method & C1 & C2 & C3 & C4 & C5 & C6 & C7 & C8 &
    		C9 & C10 & C11 & C12 & C13 & C14 & OA\\  			\hline
            {SVM}&{83.82}&{26.93}&{41.25}&{90.56} &{89.55}&{16.67}&{92.66}&{60.54}&
            {14.26}&{54.88}&{86.82}&{87.17} &{37.80}&{48.88}&{80.82}\\
            {CNN}&{97.06}&{99.79}&{92.54}&{95.01} &{93.92}&{88.12}&{97.80}&{95.75}&
            {99.95}&{70.93}&{97.72}&{99.72} &{92.34}&{96.78}&{96.58}\\
            {CNN-MRF}&{99.34}&\bf{99.89}&{98.71} &{98.68}&{97.08}&\bf{96.81}&{99.80}&\bf{96.77}&
            \bf{99.95}&{82.09}&{98.84}&\bf{99.98}&{94.72}&{97.76}& {98.70}\\
            {CV-CNN}&\bf{99.80}&{98.30}&{98.90}&{96.10}&\bf{99.60}&{93.20}&\bf{99.90}&{90.70}&
            {99.00}&\bf{98.20}&{97.00}&{99.80}&{96.60}&\bf{99.20}&{99.00}\\
            {Ours}&{99.69}&{96.35}&\bf{99.43}&\bf{98.93}&\bf{99.60}&{95.69}&{99.67}&{93.51}&		 {98.47}&{92.74}&\bf{99.64}&{99.47}&\bf{97.74}&{96.71}&\bf{99.07}\\            \hline
            \hline
		\end{tabular}
\vspace{-0.1cm}
	\end{center}
\label{tbl.Flevo}
\end{table*}

\begin{figure*}[htb]
\begin{center}
\includegraphics[height=2.7cm,width=18cm]{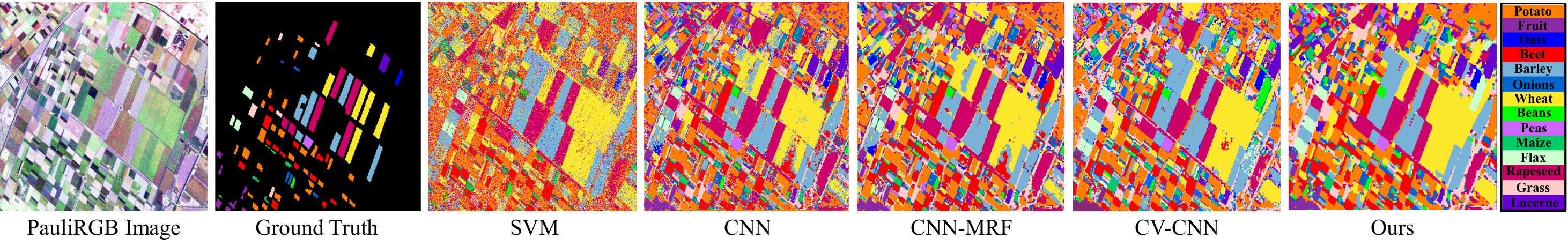}
\vspace{-0.2cm}
\caption{\small Experimental data and results of Flevoland area data set.}
\vspace{-0.4cm}
\label{fig.cmpFlevo}
\end{center}
\end{figure*}

\begin{figure*}[htb]
\begin{center}
\includegraphics[height=2.9cm,width=18cm]{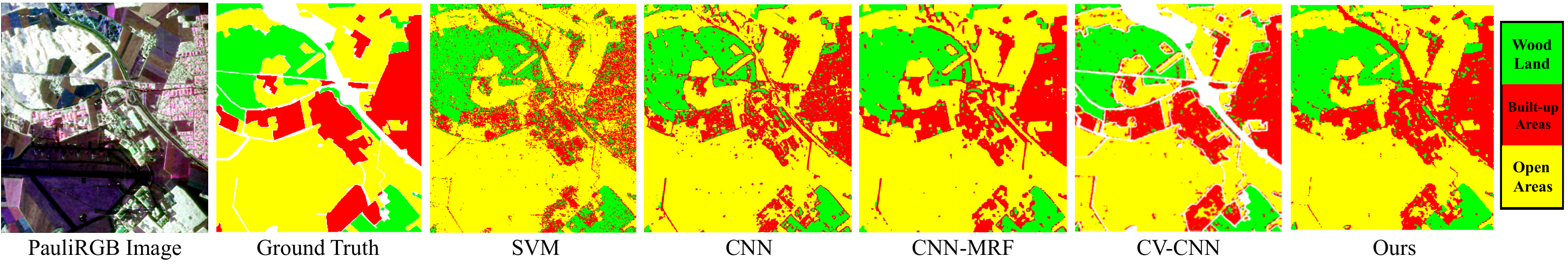}
\vspace{-0.2cm}
\caption{\small Experimental data and results of Oberpfaffenhofen area data set.}
\vspace{-0.4cm}
\label{fig.cmpOberf}
\end{center}
\end{figure*}

\subsection{Experimental Data and Experimental Settings}

Figure 2 and 3 display the experimental images employed for evaluation.
The first data set is an L-band image
collected by AIRSAR over Flevoland area in 1991 with size 1020$\times$1024.
The second data set is an E-SAR L-band image provided
by German Aerospace Center over Oberpfaffenhofen area in Germany
with size 1300$\times$1200. It is noteworthy that, for both data sets, we utilize much less labeled data than compared methods,
which are summarized in Table~\ref{tbl.usedlabels}.

In our experiments, we empirically set patch size $s$ as 7.
We performed experiments to assess the impact
of different rank values $\{2, 3, 4, 5\}$
on the classification accuracy over the Flevoland area dataset.
Experimental results show that $r=2$ achieves the best performance. Therefore, rank $r$ is set to 2 throughout the experiments.
An automatic $K$ estimation method is employed
to determine the number of Gaussian noise components.
The EM iterations stop when the iteration reaches 100
or the difference between two consecutive $\textbf{U}_s$ is smaller than 0.01.
For the CNN architecture,
the first convolutional layer contains 20 filters
with size $3\times 3$, while the second convolutional layer
involves 50 filters with size $2\times 2$.
The kernel size of the two max-pooling layer is $2\times 2$.
The first fully connected layer contains 500 units,
and the unit number of the second one takes the value of the class number.
We empirically set the learning rate $\tau$,
weight decay and momentum parameter
to 0.001, 0.0005 and 0.9, respectively.
The batch size is set as 50,
and early stopping criteria is utilized
to prevent overfitting by stopping training when the
validation accuracy continuously decreases.
The smoothness factor $\alpha_s$ is set as 5.

\subsection{Results and Comparisons}

\subsubsection{Flevoland Area Data set}

The classification results of the five compared methods on the Flevoland area data set are illustrated in Fig.~\ref{fig.cmpFlevo}.
Table~\ref{tbl.Flevo} presents the classification accuracy (\emph{CA}) and overall CA (\emph{OA}) values. Observing the figure and table, we highlight below main observations:

(a) The class label map of our proposed method presents a desirable visual effect
with preferable contextual consistency and clear label boundaries
while well preserving the image details, which is due to
noise removal of RLRMF feature representation,
feature spatial consistency encouragement of CNN,
and enforcement of label smoothness of MRF.

(b) Our method achieves the best OA with only 2\% labels used, compared with SVM, CNN and CNN-MRF using 10\% labels, and CV-CNN using 8.8\% labels.
This is attributed to the stronger feature representation and classification performance of our method, which greatly alleviates the reliance of deep learning (DL) method on large labeled sample set.

\subsubsection{Oberpfaffenhofen Area Data set}

\begin{table}[!t]
\begin{center}
\caption{CAs (\%) and OAs (\%) of Oberpfaffenhofen area data with different methods}
\vspace{-0.1cm}
\begin{tabular} {c|ccc|c} \hline \hline
{Method} & C1 & C2 & C3 & OA\\ \hline
{SVM} & {60.15} & {86.59} & {94.87} & {84.67}\\
{CNN} & {76.09} & {86.79} & {97.42} & {90.07}\\
{CNN-MRF} & {82.85} & {90.42} & {97.65} & {92.84} \\
{CV-CNN} & \bf{91.30} & {92.20} & {94.60} & {93.40}\\
{Ours} & {86.09} & \bf{95.68} & \bf{97.95}& \bf{94.58}\\
\hline \hline
\end{tabular}
\vspace{-0.5cm}
\label{tbl.Oberf}
\end{center}
\end{table}

\begin{table*}[htb]
	\caption{CAs (\%) and OAs (\%) of Flevoland area data set for ablation study}
	\vspace{-0.1cm}
	\begin{center}
		\begin{tabular}{c|cccccccccccccc|c}
			\hline	
			\hline
    		Method & C1 & C2 & C3 & C4 & C5 & C6 & C7 & C8 & C9 & C10 & C11 & C12 & C13 & C14 & OA\\
			\hline			
{RF-Raw}&{85.76}&{13.26}&{36.59}&{91.07}&{88.51}&{6.10}&{92.66}&{34.10}&
			{6.99}&{58.06}&{82.66}&{86.73}&{17.55}&{32.83}&{78.83}\\
{RF-Lee}&{98.23}&{77.14}&{0}&{98.76}&{97.38}&{0.28}&{99.24}&{91.50}&
			{31.90}&{20.39}&{95.54}&{97.90}&{66.56}&{73.54}&{91.55}\\
{RF-RLRMF}&{96.54}&{93.29}&{82.93} &{96.01}&{97.77}&{33.33}&{97.67}&{90.76}&
			{95.83}&{87.83}&{96.21}&{96.30}&{89.27}&{84.18}&{95.00}\\
{CNN-Lee}&{99.45}&\bf{99.82}&{99.00} &{94.66}&{99.25}&{53.85}&{99.65}&\bf{97.60}&
			\bf{98.81}&\bf{97.91}&{98.30}&{98.86}&{93.65}&{89.46}&{97.42}\\
{CNN-RLRMF}&{99.19}&{95.73}&{98.06} &{98.13}&{99.34}&{88.77}&{99.29}&{88.72}&
			{98.43}&{87.88}&{98.98}&{99.16}&{96.74}&{95.43}&{98.56}\\			
{Ours}&\bf{99.69}&{96.35}&\bf{99.43}&\bf{98.93}&\bf{99.60}&\bf{95.69}&\bf{99.67}&{93.51}&		 {98.47}&{92.74}&\bf{99.64}&\bf{99.47}&\bf{97.74}&\bf{96.71}&\bf{99.07}\\   	
\hline	\hline	
    \end{tabular}
	\end{center}
	\vspace{-0.1cm}
\label{tbl.ablstu}
\end{table*}

Figure~\ref{fig.cmpOberf} shows the classification results of Oberpfaffenhofen data set with five competing methods, and the numerical results are displayed in Table~\ref{tbl.Oberf}. Since the original paper of CV-CNN~\cite{zhang2017complex} did not provide the classification result of the whole image but gave the result overlaid with the ground truth labels instead, there are some void areas in the CV-CNN result. We can find from the results that, due to the combinatorial effect of RLRMF features, data augmented CNN and MRF optimization, CNN-RLRMF-MRF yields the highest OA value using less labels (0.5\%), compared with
SVM, CNN and CNN-MRF using 10\% labels, and CV-CNN using 1\% labels.

\subsection{Ablation Study}

To validate the effectiveness of the three components of the proposed method, i.e., RLRMF features, CNN and MRF optimization, we conduct following six groups of experiments
-- (1) RF-Raw: using raw features and Random forest (RF) classifier; (2) RF-Lee: using features with refined Lee noise reduction \cite{lee1999polarimetric} and RF classifier;
(3) RF-RLRMF: using RLRMF features and RF classifier; (4) CNN-Lee: using features with refined Lee noise reduction \cite{lee1999polarimetric} and CNN classifier; (5) CNN-RLRMF: using RLRMF features and CNN; (6) Ours: using RLRMF features, CNN and MRF prior-based optimization. 2\% of the ground truth labels are used in training.

\begin{figure}[!t]
\begin{center}
\includegraphics[height=5.4cm,width=7.8cm]{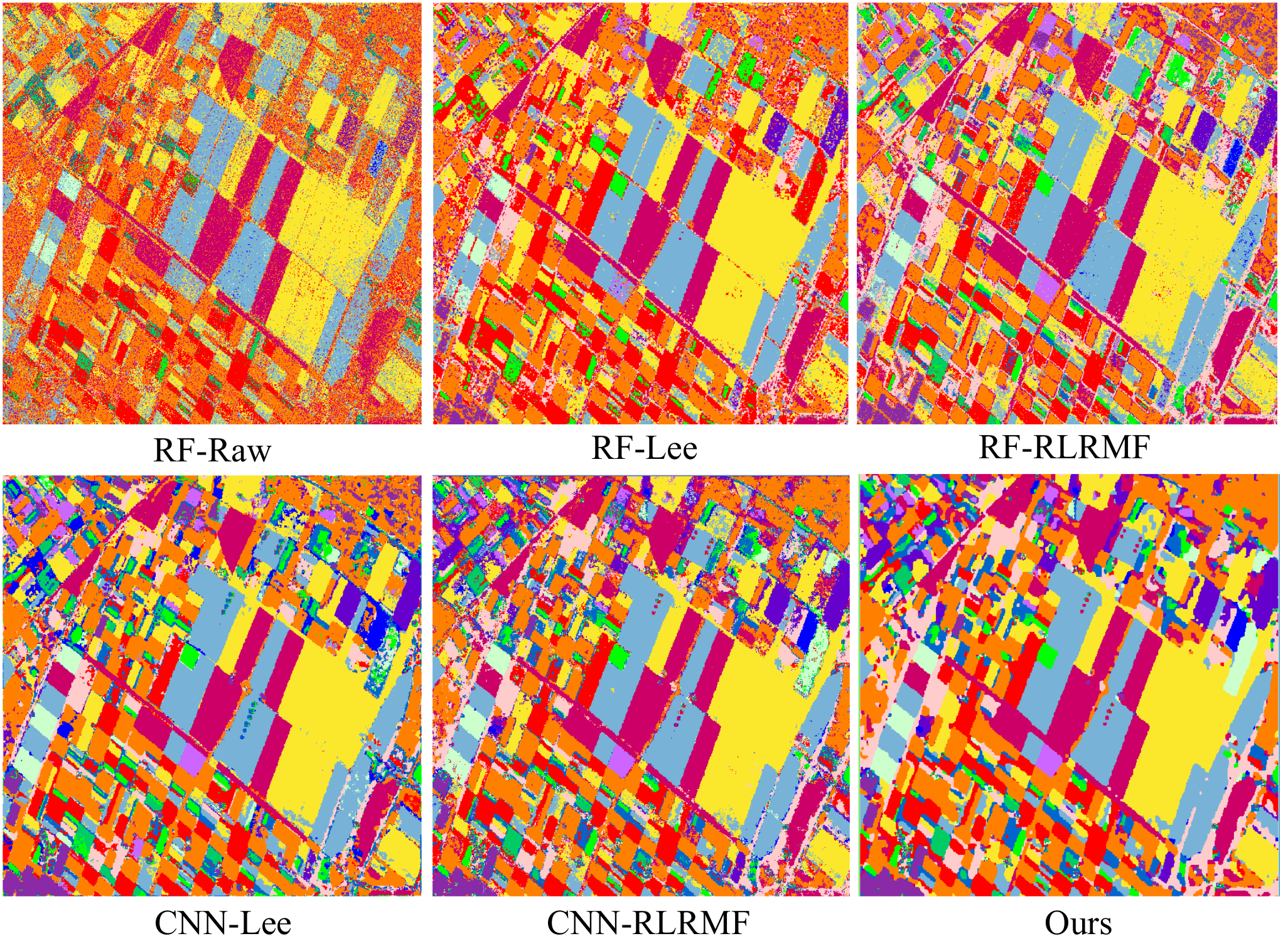}
\vspace{-0.4cm}
\caption{Experimental results for ablation study.}
\vspace{-0.4cm}
\label{fig.ablstu}
\end{center}
\end{figure}

Taking the Flevoland area data set for example, Fig.~\ref{fig.ablstu} presents the class label maps and Table~\ref{tbl.ablstu} shows the \emph{CA} values of the six compared methods, where the best results are highlighted in bold. From the figure and table, we can conclude:

(1) The class label map of RF-Raw method exhibits clear speckle-like appearance,
while this issue is obviously relieved in the result of RF-RLRMF method,
which gracefully demonstrates the noise removal effect of RLRMF features.
The RLRMF features improve the OA by 16.17 percentage points.
The RF-RLRMF and CNN-RLRMF outperform RF-Lee and CNN-Lee
by 3.45 and 1.14 percentage points, respectively, which
demonstrates the advantage of RLRMF features.

(2) CNN and MRF further promote the classification performance
and spatial consistency, which bring gains of 3.56 and 0.51 percentage points in OA respectively. Furthermore, the label boundaries
are better aligned with the image edges after MRF optimization.

\section{Conclusions}
In this paper, we presented a PolSAR classification method
involving RLRMF-based feature extraction
and MRF-based label optimization.
The advantages of our work lie in two points:
(1) The proposed method simultaneously extracts robust features and
removes noises via the MoG-based low-rank modeling.
(2) The low-rank feature extraction, CNN, and MRF effectively
improve the classification performance,
mitigating the reliance of DL method on massive labels
and promoting the spatial smoothness.
In the future, we are interested in developing
a DL-based multi-level feature fusion method \cite{hong2020more}
to suppress noise and investigating the inner relationship between
the coherency matrix elements to further boost the PolSAR image classification performance.

\bibliographystyle{ieeetr}
\bibliography{references}

\end{document}